# USING ROUGH SET AND SUPPORT VECTOR MACHINE FOR NETWORK INTRUSION DETECTION

Rung-Ching Chen [1], Kai-Fan Cheng [2] and Chia-Fen Hsieh [3]

[1] Department of Information Management Chaoyang University of Technology
Taichung Country, Taiwan, R.O.C
crching@cyut.edu.tw

[2] Department of Information Management Chaoyang University of Technology
Taichung Country, Taiwan, R.O.C
s9614627@cyut.edu.tw

[2] Department of Information Management Chaoyang University of Technology
Taichung Country, Taiwan, R.O.C
s9733901@cyut.edu.tw

## ABSTRACT

*The main function of IDS (Intrusion Detection System) is to protect the system, analyze and predict the behaviors of users. Then these behaviors will be considered an attack or a normal behavior. Though IDS has been developed for many years, the large number of return alert messages makes managers maintain system inefficiently. In this paper, we use RST (Rough Set Theory) and SVM (Support Vector Machine) to detect intrusions. First, RST is used to preprocess the data and reduce the dimensions. Next, the features were selected by RST will be sent to SVM model to learn and test respectively. The method is effective to decrease the space density of data. The experiments will compare the results with different methods and show RST and SVM schema could improve the false positive rate and accuracy.*

## 1. INTRODUCTION

The intrusion behaviours cause the great damage of systems. So enterprises search for intrusion detection systems to protect their systems. The traditional technology such as firewall is used to defense attacks. Thus, the IDS (Intrusion Detection System) is usually used to enhance the network security of enterprises.

The major difference between firewall and IDS system is that firewall is a manual passive defense system. Comparatively, IDS could collect packets online from the network. After collecting them, IDS will monitor and analyze these packets. So, IDS system acts as the "second line of defense". Finally, it will provide the detecting results for managers. The detecting results could be either attack or normal behaviour. An ideal IDS system has a 100% attack detection rate along with a 0% false positive rate, but it is hard to achieve. Detecting illegal behaviours of the host or network is the major object of IDS. The IDS is actually such a system to detect some illegal behaviour. One of the ability of IDS is it could monitor various activities on the network. IDS will send a warning message to the managers if it detects an attack. Briefly, the aim of intrusion detection is to identify malicious attacks. There are two main methods of IDS: misuse and anomaly [11]. The idea of misuse detection is to establish a pattern or a signature form so that the same attack can be detected. We will describe it in next section. The other idea here is to establish a normal activity profile for system. Anomaly detection utilizes soft computing methods to detect attacks, for instance, Neural network [10], Static analysis, Data mining [8] etc. Anomaly detection can detect new attacks, but it has a higher false positive rate. Large number of transmission packages will lead to computation overloading and worse performance. In this paper, we will purpose an intrusion detection method to reduce the features of transferring packages using the method of Rough Set Theory. If the number of the key features is decreased correctly, the noise of data to affect the systems analysis performance will be lower. Primary experiment results prove our research can improve the attack detection rate.

The remainders of the paper are organized as follows. Section 2 presents the related literature of intrusion detection systems. Section 3 introduces our methodology. Section 4 shows the experiments results and Section 5 is conclusions and future works.

## 2. RELATED WORKS

### 2.1. The type intrusion methods

The IDS improves the attack detection rate (ADR) and decrease the false alarm rate (FAR) [7][8][14]. The Denial-of service (DOS) attacks also called the Distribution Denial-of Service (DDOS) attack is shown in Figure 1. The attacker uses large numbers of computers to login system in a short period of time or to transfer mass numbers of packages. These actions will lead to overloading of the host. The network services of host will stop. Large resource of the host such as the utilization of CPU and memory and the networks bandwidth are consumed by the attack. SYN-flood, Smurf, Teardrop and Ping of Death are belonging to the DoS attack.

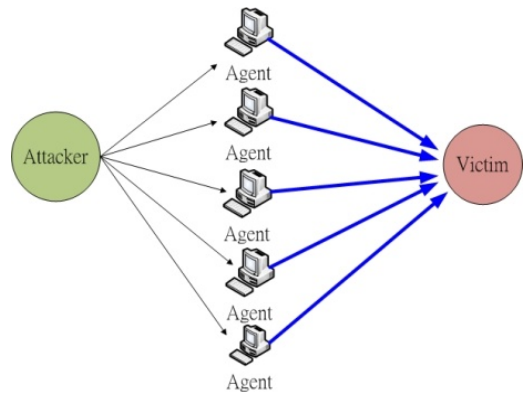

Figure 1 The schema of DDoS/DoS.

The intruders find the user's account or the system setting out of gear. The R2L attack tries to intrusion the system. SQL Injection is one kind of the R2L attacks. U2R attack uses the unauthorized account to control system. The methods of U2R use the virus, worm, to overload buffer of memory to exceeding the limitation of memory access. For instance, the several of Buffer, Overflow Attacks are typical traditional U2R attack. Hackers scan the protocol of the computer before they launch attacks by port Scan action. The hackers find the weak spots and software design weakness to intrude a system. Port Scanning and the Ping-Sweep are typical traditional port scan attack.

### 2.2. Intrusion detection system

Intrusion Detection System has misuse detection and anomaly detection(Figure 2). The known attack behaviours are constructed from misuse detection attribute database in the development stage. Misuse detection system compares user behaviours with attribute database to find intrusions. Anomaly detection system defines system exactly normal behaviours in rule database. The contrast between collects system parameter and defines behaviours can find the misbehaviours in system.

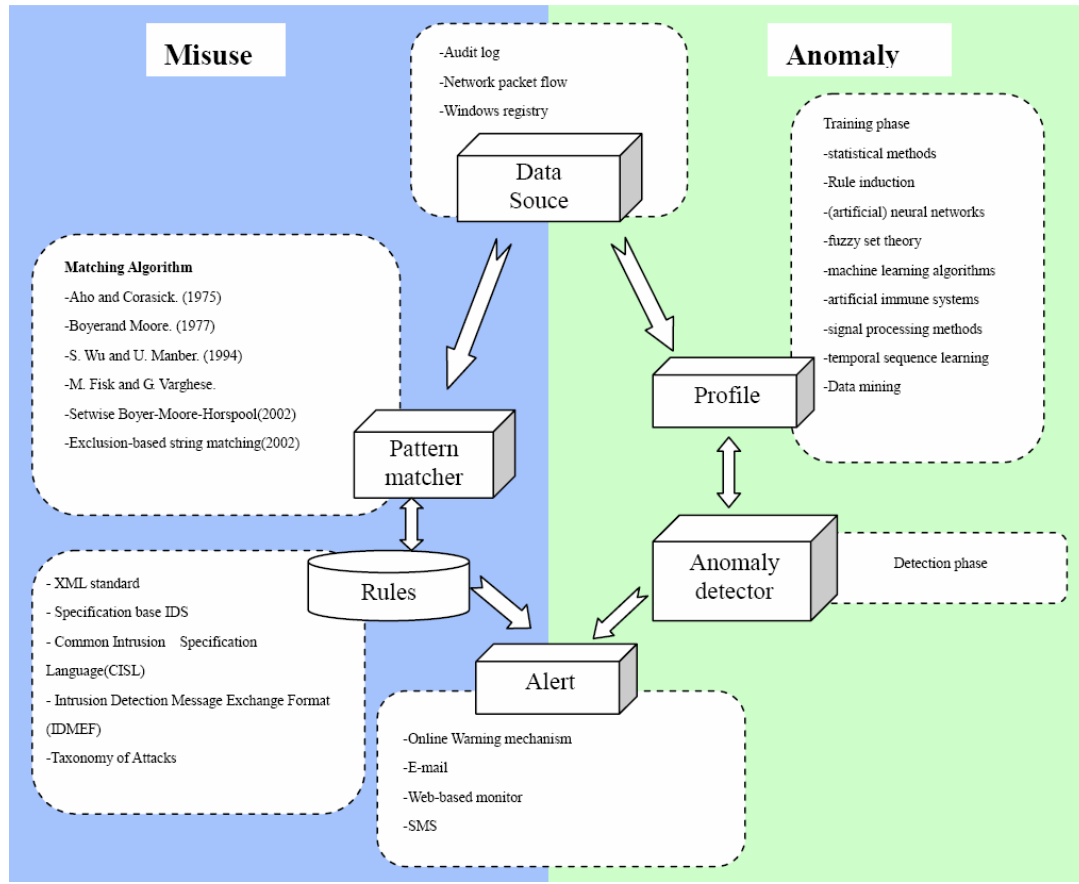

Figure 2 The Flow Chart of Misuse Detection and Anomaly Detection Application Flow Chart

### A. Misuse Detection

Misuse detection sets up the attack behaviours based on known attack behaviours during the development stage. The misuse detection is similar to antivirus software. The antivirus software compares the scanned data with known virus code. If system finds un-normal attributes, the virus is existence and removes it. Hence, misuse detection collects the known attack behaviours from attribute database. If the attack behaviour is similar to the one in database, the misuse detection can defend it before the intruder destroys our system.

### B. Anomaly Detection

Anomaly detection is different from misuse detection. The system constructs user model based on normal users have behaviours. When user has misbehaviours, the system notifies users that has an intruder. The main drawback of anomaly detection is that the detection is depended on the latest attack models, so it can't indentify new attack

International Journal of Network Security & Its Applications (IJNSA),Vol 1, No 1, April 2009behaviours. The intruder attack methods will be changed, so anomaly detection system collects normal behaviours and detects intruding using normal behaviours. The anomaly detection system has a party with clearly defined correct user behaviours. The problem is intruder uses normal behaviours to attack the system.

IDS system monitors the packages transmissions on the network. While malice behaviours have happen, IDS will send an alert to the network manager or use a related method to defense the attacks. Most intrusion detection systems are classified as either a NIDS (Network Based Intrusion Detection System) or a HIDS (Host Based Intrusion Detection System) [6][13]. In general, NIDS is located between host and firewall. HIDS was usually installed on a server or main computer as shown in Figure 3. NIDS collects and analyzes the information at the host. NIDS could monitor the data real time on the network. If NIDS finds illegal behaviours, it will send messages to the managers. Comparatively, HIDS monitors the activities of the host. So it can determine whether an attack or not. The data of HIDS is caught by the host, so it is not easy to be influenced by some methods, just like encryption. Entropy has been used in intrusion detection for a long time. B. Balajinath et al. used entropy in learning behaviour model of intrusion detection in 2001[2]. TF-IDF is often applied to IDS, too. Such as Wun-Hwa Chen et al. compared SVM to ANN for intrusion detection, their methods are based on TF-IDF. The Unauthorized Access from a Remote Machine (R2L), and The Unauthorized Access to Local Super-user Privileges (U2R) both are intrusion behaviours which will be detected by HIDS.

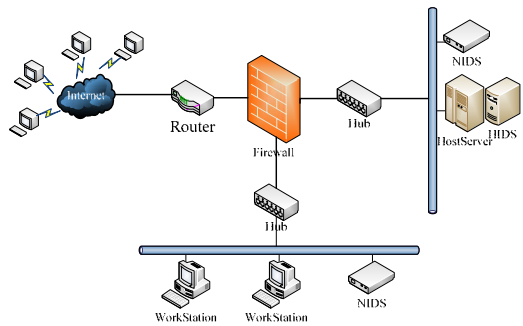

Figure 3 The hybrid system of IDS.

## 3. THE METHODOLOGY

The flowchart is our intrusion detection method shown as Figure 4 which comprises three steps. First, data pre-processing and data discretion are utilized to do data arrangement. Next, the RST is used to find useful features. Finally, the system uses the SVM to classify the data [2][12], which will be described as follows.

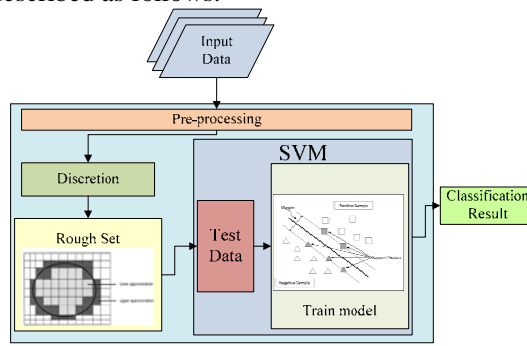

Figure 4 The workflow of the system in training phase.

International Journal of Network Security & Its Applications (IJNSA),Vol 1, No 1, April 2009## 3.1 Pre-processing

In this section, we will introduce the KDD cup'99 and use the database to test the system performance. The KDD Cup 1999 Data is original from 1998 DARPA Intrusion Detection Evaluation [9]. A process can be composed of many system calls. A system call is a text record. In this phase, some useless data will be filtered and modified. For example, some text items need to be converted into numbers. Every process in the database has 41 attributes shown in Table 1.

Table 1 KDD cup'99 features

| No. | Features | No. | Features |
|---|---|---|---|
| 1 | duration | 22 | is_guest_login |
| 2 | protocol_type | 23 | count |
| 3 | service | 24 | srv_count |
| 4 | flag | 25 | serror_rate |
| 5 | src_bytes | 26 | srv_serror_rate |
| 6 | dst_bytes | 27 | rerror_rate |
| 7 | land | 28 | srv_rerror_rate |
| 8 | wrong_fragment | 29 | same_srv_rate |
| 9 | urgent | 30 | diff_srv_rate |
| 10 | hot | 31 | srv_diff_host_rate |
| 11 | num_failed_logins | 32 | dst_host_count |
| 12 | logged_in | 33 | dst_host_srv_count |
| 13 | num_compromised | 34 | dst_host_same_srv_rate |
| 14 | root_shell | 35 | dst_host_diff_srv_rate |
| 15 | su_attempted | 36 | dst_host_same_src_port_rate |
| 16 | num_root | 37 | dst_host_srv_diff_host_rate |
| 17 | num_file_creations | 38 | dst_host_serror_rate |
| 18 | num_shells | 39 | dst_host_srv_serror_rate |
| 19 | num_access_files | 40 | dst_host_rerror_rate |
| 20 | num_outbound_cmds | 41 | dst_host_srv_rerror_rate |
| 21 | is_host_login | | |

This dataset has been used for the Third International Knowledge Discovery and Data Mining Tools Competition. The task of this competition was to build a network detector to find "bad" connections and "good" connections. For "bad" connections, the attack has categories, shown in Table 2. An example of KDD cup'99 [9] is shown in Figure 5. There are several text words in the dataset. The system will transform text into numeric values in advance. For example, the service type of "tcp" is mapping to 3 and the system will follow Table 3 to transform it into the numeric form, as described in Figure 6.

```
0,tcp,http,SF,181,5450,0,0,0,0,0,1,0,0,0,0,0,0,0,0,0,0,8,8,0.00,0.00,0.00,0.00,1.00,0.00,0
.00,9,9,1.00,0.00,0.11,0.00,0.00,0.00,0.00,0.00,normal.
0,tcp,http,SF,239,486,0,0,0,0,0,1,0,0,0,0,0,0,0,0,0,0,8,8,0.00,0.00,0.00,0.00,1.00,0.00,0.
00,19,19,1.00,0.00,0.05,0.00,0.00,0.00,0.00,0.00,normal.
0,tcp,http,SF,235,1337,0,0,0,0,0,1,0,0,0,0,0,0,0,0,0,0,8,8,0.00,0.00,0.00,0.00,1.00,0.00,0
.00,29,29,1.00,0.00,0.03,0.00,0.00,0.00,0.00,0.00,normal.
```
Figure 5 The original data of KDD cup'99

```
0,3,19,10,181,5450,0,0,0,0,0,1,0,0,0,0,0,0,0,0,0,0,8,8,0,0,0,0,1,0,0,9,9,1,0,0.11,0,0,0,0,0,0
0,3,19,10,239,486,0,0,0,0,0,1,0,0,0,0,0,0,0,0,0,0,8,8,0,0,0,0,1,0,0,19,19,1,0,0.05,0,0,0,0,
0
0,3,19,10,235,1337,0,0,0,0,0,1,0,0,0,0,0,0,0,0,0,0,8,8,0,0,0,0,1,0,0,29,29,1,0,0.03,0,0,0,
0,0
```
Figure 6 After transform from original data

Table 2 Data types and attacks classification

| Attack types | class | Attack types | class |
|---|---|---|---|
| Normal | Normal | guess_passwd | R2L |
| apacha2 | DoS | Imap | R2L |
| back | DoS | multihop | R2L |
| land | DoS | named | R2L |
| mailbomb | DoS | phf | R2L |
| netune | DoS | sendmail | R2L |
| pod | DoS | snmpgetattack | R2L |
| processtable | DoS | snmpguess | R2L |
| smurf | DoS | spy | R2L |
| teardrop | DoS | warezclient | R2L |
| udpstorm | DoS | warezmaster | R2L |
| buffer_overflow | U2R | worm | R2L |
| httprunnel | U2R | xlock | R2L |
| loadmodule | U2R | xsnoop | R2L |
| perl | U2R | Ipsweep | Probe |
| ps | U2R | mscan | Probe |
| rootkit | U2R | nmap | Probe |
| sqlattack | U2R | portsweep | Probe |
| xterm | U2R | saint | Probe |
| ftp_write | R2L | satan | Probe |

Table 3 Transformation table

| Types | Class | No. |
|---|---|---|
| Protocol_type | TCP | 3 |
| Protocol_type | UDP | 7 |
| Protocol_type | ICMP | 9 |
| Flag | OTH | 1 |
| Flag | REJ | 2 |
| Flag | RSTO | 3 |
| Flag | RSTOS0 | 4 |
| Flag | RSTR | 5 |
| Flag | S0 | 6 |
| Flag | S1 | 7 |
| Flag | S2 | 8 |
| Flag | S3 | 9 |
| Flag | SF | 10 |
| Flag | SH | 11 |
| Attack or Normal | Attack | 1 |
| Attack or Normal | Normal | 0 |

## 3.2 Feature selection by rough set

Using the RST reduces the attributes for SVM operation. Rough Set Theory [3][5][14][15] is one of data-mining methods which reduces the features from large numbers of data. Using RST needs to build the decision table or the information table. The decision table describes the features of processes. Formally, an information system IS (or an approximation space) can be shown as follows.

$$IS = (U, A) \quad (1)$$

Where U is the Universe (a dataset of process, U=$\{x_1,x_2,x_3,x_4,x_5,x_6,...,x_m\}$) and A presents the attributes of a process, for instances, (A=$\{a_1,a_2,a_3,a_4,a_5\}$). The definition of an information function is $f_a$: U→ $V_a$, $V_a$ is the set of values of the attributes. For example, the values of U and A are listed as follows and they are mapping to $V_i$.

U= $\{x_1, x_2, x_3, x_4, x_5, x_6, ..., x_m\}$
A= $\{a_1, a_2, a_3, a_4, a_5\}$
$V_1$= $\{1, 2, 3, 4\}$
$V_2$= $\{1, 2, 3, 4, 5\}$
$V_3$= $\{1, 2, 3, 4, 5\}$
$V_4$= $\{1, 2, 3\}$

For every set of attributes B⊆A, if b ($x_i$) = b ($x_j$)( every b⊆B), there is an indiscernible relation Ind(B). Continuous, to define the basic concepts, namely the Upper approximations and Lower approximations of a set let X represents the elements of subset of the universe U (X⊆U). The lower approximations of X in B (B⊆A) represents as $\overline{BX}$ such as follows.

$$\overline{BX} = \{X_i \in U \,|\, [X_i]_{ind(B)} \subset X\} \quad (2)$$

The lower approximations of set X of process $x_i$, which contained X of elementary set in the space B. The upper approximation of set X is BX. BX represents the union of the elementary which is a non-empty intersection with X.

$$BX = \{X_i \in U \,|\, [X_i]_{ind(B)} \cap X \neq 0\} \quad (3)$$

For any object xi of lower approximation of X ($x_i \in \overline{BX}$), it is certainly belongs to X. For object of xi of upper approximations of X ($x_i \in BX$), it is called a boundary of X in U. The difference of upper and lower approximations is:

$$BNP = BX - \overline{BX} \quad (4)$$

If the upper and lower approximations are identical (BX = $\overline{BX}$), the set X is definable; otherwise, set X is indefinable in U. There are four types of the set of indefinable in U. Ø represents an empty set.

- If $\overline{BX}$ ≠ Ø and BX ≠ U, the set of X represents roughly definable in U;
- If $\overline{BX}$ ≠ Ø and BX=U, the set of X represents externally indefinable in U;
- If $\overline{BX}$ = Ø and BX ≠ U, the set of X represents internally indefinable in U;
- If $\overline{BX}$ = Ø and BX ≠ U, the set of X represents totally indefinable in U.

Using all attributes to do intrusion detection is ineffective. In this paper, RST is used to combine the similar attributes and to reduce the number of attributes. So it can enhance the processing speed and to promote the detection rate for intrusion detection. An example of RST is shown in Figure 7.

```
O:1,0,3,19,10,181,5450,0,0,0,0,0,1,0,0,0,0,0,0,0,0,0,0,0,8,8,0,0,0,0,1,0,0,9,9,1,0,0.11,0,0,
0,0,0
O:2,0,3,19,10,239,486,0,0,0,0,0,1,0,0,0,0,0,0,0,0,0,0,0,8,8,0,0,0,0,1,0,0,19,19,1,0,0.05,0,0
,0,0,0
O:3,0,3,19,10,235,1337,0,0,0,0,0,1,0,0,0,0,0,0,0,0,0,0,0,8,8,0,0,0,0,1,0,0,29,29,1,0,0.03,0,
0,0,0,0
```
Figure 7 The Rough set input data form

### 3.3 Intrusion estimation

In this paper, we construct an SVM [7] model for classification. While intrusion behaviors happens. SVM will detect the intrusion. SVM uses a high dimension space to find a hyper-plane to perform binary classification, where the error rate is minimal. The SVM can handle the problem of linear inseparability. The SVM uses a portion of the data to train the system, finding several support vectors that represent the training data. These support vectors will be formed a model by the SVM, representing a category. According to this model, the SVM will classify a given unknown document. A basic input data format and output data domains are listed as follows.

$(x_i, y_i),…,(x_n, y_n), x \in R^m, y \in \{+1,-1\}$ (5)

Where $(x_i, y_i),…,(x_n, y_n)$ are a train data, n is the numbers of samples, m is the inputs vector, and y belongs to category of +1 or -1 respectively. On the problem of linear, a hyper plan can divided into the two categories as shown in Figure 8. The hyper plan formula is:

$(w \cdot x) + b = 0$ (6)

The category formula is:

$(w \cdot x) + b \geq if\ y_i = +1$ (7)

$(w \cdot x) + b \leq if\ y_i = -1$ (8)

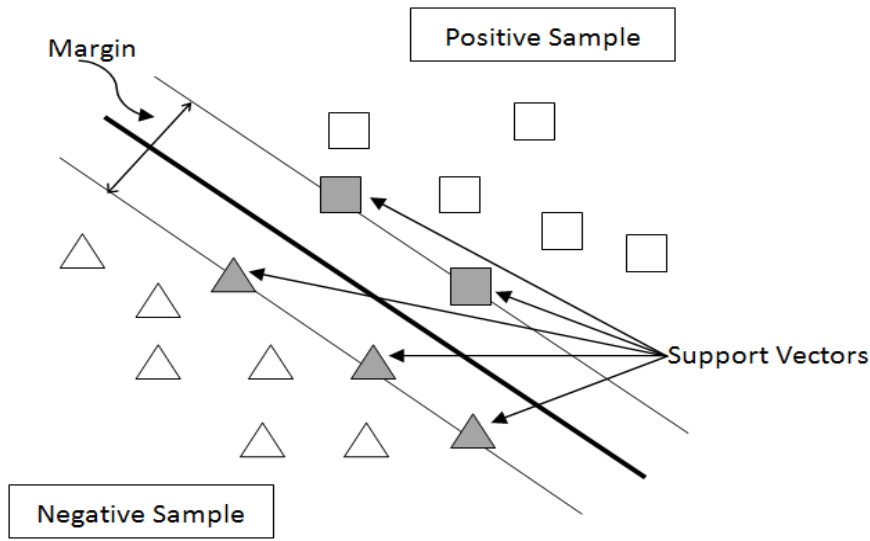

Figure 8 The hyper-plane of SVM.

However, for many problems they are not easy to find hyper planes to classify the data. The SVM has several kernel functions that users can apply to solving different problems. Selecting the appropriate kernel function can solve the problem of linear inseparability. Another important capability of the SVM is that it can deal with linear inseparable problems. Internal product operations affect the classification function. A suitable inner product function K $(X_i \cdot X_j)$ can solve certain linear inseparable problems without increasing the complexity of the calculation.

## 4. EXPERIMENTS AND DISCUSSIONS

The experiment used a AMD Athlon™ 64 X2 Dual Core Processor 5000+ 2.59G MHz computer with 512MB RAM, and implemented on a Windows XP Professional operating system. The data was collected from the MIT Lincoln Lab, 1998 DARPA intrusion detection evaluation program. The data set offers seven weeks of training data and two weeks of testing data. We randomly selected some of the data for training data and testing data. The DARPA data is labeled with a session number, each session including several processes, which in turn consist of system calls. A total of 24,701 processes were used as training data, while 15,551 processes were used as testing data that are shown in Table 4. We have 41 features initially, which are presented and calculated their frequency in this database. LibSVM was used as our classification tool [3].

The total of 41 features, entropy and 29 RST feature values are used to train the three SVM models. After the pre-processing process, data are formatted for RST. Then, the data of discretion is imported the RST by analysis tool. RST upper and lower approximation is utilized to seek the relation and un-relation of feature attributes. The features were selected by RST which are shown in Table 5.

Table 4. The training data and test data

|        | Train | Ratio   | Test  | Ratio   |
|--------|-------|---------|-------|---------|
| Normal | 4863  | 19.69%  | 3029  | 19.48%  |
| Probe  | 205   | 0.83%   | 208   | 1.34%   |
| DoS    | 19572 | 79.24%  | 11492 | 73.90%  |
| U2R    | 2     | 0.01%   | 11    | 0.07%   |
| R2L    | 56    | 0.23%   | 809   | 5.20%   |
| Total  | 24701 | 100.00% | 15551 | 100.00% |

Protocol type, Services, and Flag are transformed into numerical values. The LIBSVM data format is:

LIBSVM format: [Label] 1: $f_1 f_2 f_3 f_4 ... f_n$. (9)

Before using SVM classification [16], we need to do "scaling". This action is to increase the accuracy, decrease the overlap, and reduce complexity [7]. Our system uses the kernel of RBF $K(x, y) = \lambda^{-\gamma \|x-y\|^2}$.

In the SVM classify, we will take the tool to do classification the attributes of features before the pre-processing step. We use the SVM to train and to test processes. Base on the estimation enhance operate in the intrusion detections, the estimation formula is listed as follows. The output of SVM is 1 or -1. If the output is 1, it is an intrusion behaviour on the model. If the output is -1, it is normal. Using LibSVM tool [7] to classification estimate intrusion or not.

Table 5. The features after RST operation

| No. | Features | No. | Features |
|---|---|---|---|
| 1 | Duration | 16 | rerror_rate |
| 2 | protocol_type | 17 | same_srv_rate |
| 3 | src_bytes | 18 | diff_srv_rate |
| 4 | dst_bytes | 19 | srv_diff_host_rate |
| 5 | wrong_fragment | 20 | dst_host_count |
| 6 | num_failed_logins | 21 | dst_host_srv_count |
| 7 | logged_in | 22 | dst_host_same_srv_rate |
| 8 | num_compromised | 23 | dst_host_diff_srv_rate |
| 9 | root_shell | 24 | dst_host_same_src_port_rate |
| 10 | num_root | 25 | dst_host_srv_diff_host_rate |
| 11 | num_file_creations | 26 | dst_host_serror_rate |
| 12 | num_shells | 27 | dst_host_srv_serror_rate |
| 13 | num_access_files | 28 | dst_host_rerror_rate |
| 14 | Count | 29 | dst_host_srv_rerror_rate |
| 15 | serror_rate | | |

## 4.1 The Estimation Method of Wired Networks IDS

In wired environment, the resource and the energy of a computer is infinite. The intruder intrudes operation system by virus or worms, such as Trojan horse. Hence, the IDS trains intruder behaviors model by an artificial intelligence (AI) method. The system trains old data for new attack behaviors. The main estimating methods are precision and recall. The wired network IDS estimates parameters shown in Table 6.

Table 6 The Wired Network IDS Estimation

| Parameter | Definition |
|---|---|
| True Positive Rate (TP) | Attack occur and alarm raised |
| False Positive Rate (FP) | No attack but alarm raised |
| True Negative Rate (TN) | No attack and no alarm |
| False Negative Rate (FN) | Attack occur but no alarm |

Helmer et al (2002) [4] shows the precision formula in (10). Precision stands for that the attack has been occurred and the IDS detects correctly. This formula use TP to divide TP adds FP to find precision rate.

$$\text{Precision} = \frac{TP}{TP+FP} \tag{10}$$

Recall stands for an attack is happened and IDS detects attacks from really attacks. This formula use TP divide TP adds FN to find recall value shown in formula (11)

$$\text{Recall} = \frac{TP}{TP+FN} \tag{11}$$

Agarwal et al (2002) [1] observes that overall formula. When the attack has been occurred, IDS detects attacks correctly. This formula use TP adds TN to divide TP adds FP adds FN adds TN to estimate overall accuracy which is shown in formula (12).

$$\text{Overall} = \frac{TP+TN}{TP+FP+FN+TN} \quad (12)$$

The false alarm is defined as an intrusion has been occurred but IDS can not detect correctly or intrusion happens. The formula uses FP adds FN to divide TP adds FP adds FN adds TN to calculate false alarm rate which is shown in formula (13).

$$\text{False Alarm} = \frac{FP+FN}{TP+FP+FN+TN} \quad (13)$$

Those formulas mainly use to estimate the effects of wired network IDS.

The research object is to increase the accuracy of SVM. We use the RST to reduce features. SVM use to archive the supervisor learning and find a suitable kernel of SVM classification. Using the test data evaluates system performance. The data distribution of training data and testing data are shown in Table 6. User can use this model to evaluate the IDS. We use (1) all 41 features, (2) entropy features and 29 RST features values to train three SVM models respectively, and to compare the accuracy of the three models. To estimate the performance of the system, three important formulas are used to evaluate system accuracy [12]; attack detection rate (ADR), false positive rate (FPR) and system accuracy (SA).

$$\text{Attack Detection Rate} = \frac{\text{Total number of attacks}}{\text{Total number of detected attacks}} \times 100\% \quad (14)$$

$$\text{False Positive Rate} = \frac{\text{Total number of misclassified processes}}{\text{Total number of normal processes}} \times 100\% \quad (15)$$

$$\text{Accuracy Rate} = \frac{\text{Total number of correct classified processes}}{\text{Total number of processes}} \times 100\% \quad (16)$$

The experiments tested attack detection rate, false positive rate and accuracy among 41 features input SVM, Entropy inputs SVM, and after Rough Set reduces feature input SVM. Accuracy of 41 features input to SVM is 86.79% but the ADR is only 70.03%. The result of this study is not well because the feature is not reduction of attributes and the loading is too heavy for the system. The Entropy features input SVM has using reduction of feature and the ADR reaches 92.44% but the accuracy is only 73.83%. The accuracy of our proposed method is the best but its false position rate and attack detection rate are worse than Entropy to SVM. Table 6 shows the results.

Table 6. Comparison of three methods using SVM

|  | Attack Detection Rate | False Positive Rate | Accuracy |
|---|---|---|---|
| 41 features to SVM | 70.03% | 29.97% | 86.79% |
| Entropy to SVM | 92.44% | 7.56% | 73.83% |
| Rough Set of SVM | 86.72% | 13.27% | 89.13% |

## 5. CONCLUSIONS AND FUTURE WORKS

In this paper, we have proposed an intrusion detection method using an SVM based system on a RST to reduce the number of features from 41 to 29. We also compared the performance of the SVM with that of a full features and Entropy. Our framework RST-SVM method result has a higher accuracy as compared to either full feature or entropy. The experiment demonstrates that RST-SVM yields a better accuracy.

In the future, we will increase number of testing data for our system and to find vary of accuracy. We also hope to combine RST method and genetic algorithm to improve the accuracy of IDS.

**Authors**

**Rung-Ching Chen** received the B.S. degree from department of electrical engineering in 1987, and the M. S. degree from the institute of computer engineering in 1990, both from National Taiwan University of Science and Technology, Taipei, Taiwan. In 1998, he received the Ph.D. degree from the department of applied mathematics in computer science sessions, National Chung Tsing University. He is Dean of College of Informatics in Chaoyang University of Technology and he is now a professor at the Department of Information Management in Chaoyang University of Technology, Taichung, Taiwan. His research interests include web technology, pattern recognition, and applied soft computing and network security.

**Kai-Fan Cheng** received the University degree from the department of Information Management of the Overseas Chinese College of Commerce, Taichung, Taiwan, in 2007. He is a Graduate student from the department of Information Management at Chaoyang University of Technology, Taichung, Taiwan. His research interests include the security issue (intrusion detection in particular) of ad hoc networks and wired networks.

**Chia-Fen Hsieh** received the Graduate degree from department of Information Management at Chaoyang University of Technology in 2008. He is a candidate for doctor's degree from graduate institute of informatics at Chaoyang University of Technology, Taichung, Taiwan. His research interests include the security issue (intrusion detection in particular) of wireless sensor networks, ad hoc networks and wired networks.